\documentclass{article}

\PassOptionsToPackage{numbers, compress}{natbib}

\usepackage[final]{neurips_2024_ml4ps}

\usepackage[utf8]{inputenc} 
\usepackage[T1]{fontenc}    
\usepackage{hyperref}       
\usepackage{url}            
\usepackage{booktabs}       
\usepackage{amsfonts}       
\usepackage{nicefrac}       
\usepackage{microtype}      
\usepackage{xcolor}         
\usepackage{amsmath}
\usepackage{listings}
\usepackage{graphicx}

\usepackage{listings}
\usepackage{siunitx}
\usepackage{xcolor}
\definecolor{codegreen}{rgb}{0,0.6,0}
\definecolor{codegray}{rgb}{0.5,0.5,0.5}
\definecolor{codepurple}{rgb}{0.58,0,0.82}
\definecolor{backcolour}{rgb}{0.95,0.95,0.92}
\usepackage{multirow}
\usepackage{caption}
\usepackage{subcaption}
\usepackage{booktabs,xcolor,colortbl}
\usepackage{xparse}
\ExplSyntaxOn
\NewDocumentCommand{\longdash}{ O{2} }
 {
  --\prg_replicate:nn { #1 - 1 } { \negthinspace -- }
 }
\ExplSyntaxOff

\lstdefinestyle{mystyle}{
    backgroundcolor=\color{backcolour}, 
    commentstyle=\color{codegreen},
    keywordstyle=\color{magenta},
    numberstyle=\tiny\color{codegray},
    stringstyle=\color{codepurple},
    basicstyle=\ttfamily\footnotesize,
    breakatwhitespace=false,         
    breaklines=true,                 
    captionpos=b,                    
    keepspaces=true,                 
    numbers=left,                    
    numbersep=5pt,                  
    showspaces=false,                
    showstringspaces=false,
    showtabs=false,                  
    tabsize=2,
    escapeinside={(*@}{@*)}
}

\title{Open-Source Molecular Processing Pipeline for Generating Molecules}

\author{V Shreyas$^{1,2}$ \quad Jose Siguenza$^{2}$ \quad Karan Bania$^{1,2}$ \quad Bharath Ramsundar$^{2}$ \\
$^1$BITS Pilani Goa Campus \quad $^2$Deep Forest Sciences \\
\texttt{\{shreyas, bharath\}@deepforestsci.com}\\
}

\begin{document}

\maketitle

\begin{abstract}
  Generative models for molecules have shown considerable promise for use in computational chemistry, but remain difficult to use for non-experts. 
  For this reason, we introduce open-source infrastructure for easily building generative molecular models into the 
  widely used DeepChem~\citep{deepchem} library with the aim of creating a robust and reusable molecular generation pipeline. 
  In particular, we add high quality PyTorch~\citep{pytorch} implementations of the Molecular Generative Adversarial Networks (MolGAN)~\citep{molgan} and Normalizing Flows~\citep{nflows}. 
  Our implementations show strong performance comparable with past work~\citep{molgrow, molgan}.
\end{abstract}

\section{Introduction}
\label{Introduction}
The discovery of new molecules and materials is key to addressing challenges in chemistry, such as treating diseases and combating climate change~\citep{retro-synthesis, main-paper}. 
Traditional methods, however, are time-consuming and costly, limiting the exploration of the vast chemical space~\citep{drug-chemical-space}. 
Generative models offer a deep learning-based solution, designing molecules with desired properties more efficiently. Despite their potential, these models typically require significant expertise in Python and machine learning.

To address this, we introduce open-source implementations of Molecular Generative Adversarial Networks (MolGAN)~\citep{molgan} and Normalizing Flow models~\citep{nflows} in pytorch into DeepChem~\citep{deepchem}, a widely used molecular machine learning library. 
MolGAN utilizes adversarial training~\citep{gan} to generate novel molecules, while Normalizing Flow models employ exact likelihood methods for molecular generation. 
Our contributions simplify their use by providing accessible pipelines that require minimal prior knowledge, while also allowing advanced users to modify the models as needed.

\section{Methods and Background}
\subsection{DeepChem and Generative Molecular Models} 
\label{deepchem_gmm} DeepChem~\citep{deepchem} is a versatile open-source Python library tailored for machine learning on molecular and quantum datasets~\citep{deepchem_alt}. Its framework supports applications in areas such as drug discovery and biotech~\citep{moleculenet}, breaking down scientific tasks into workflows built from core primitives. 
DeepChem has facilitated significant advancements, including large-scale molecular machine learning benchmarks via MoleculeNet~\citep{moleculenet}, protein-ligand modeling~\citep{gomes2017atomic}, and generative molecule modeling~\citep{fastflow}.

While older DeepChem implementations of MolGANs and Normalizing Flows used TensorFlow, we migrate these models to PyTorch, ensuring tighter integration with DeepChem’s ecosystem and broader compatibility. This enables users to leverage DeepChem's extensive layer library to experiment and build new models. 

\subsection{Representation of Molecules}
The strength of neural networks lies in their ability to take in a complex input representation and transform it into a latent representation needed to solve a particular task. 
In this way, the choice of input representation plays a key role in governing how the model learns information about the molecule. 
Input representations often fall into one of two categories: (1) one-dimensional (e.g., string-based representations), (2) two-dimensional (e.g., molecular graphs).

\subsubsection{One-dimensional representations}
The most common one-dimensional representation of molecules is SMILES (Simplified Molecular Input Line Entry System)~\citep{smiles}, which transforms a molecule into a sequence of characters based on predefined atom ordering rules. 
This representation enables the use of neural network architectures developed for language processing. 
For instance, previous work~\citep{rgb_ACD, oli_denovo} used recurrent neural networks as generative models to create SMILES strings. 
However, these methods often produce invalid SMILES that cannot be converted to molecular structures due to their disregard for SMILES grammar. 
To overcome this limitation,~\citep{selfies} introduced SELFIES (Self-Referencing Embedded Strings), an improved string representation. 
With SELFIES, a recurrent neural network can generate molecules with 100\% validity, though validity here pertains to valency rules and does not guarantee molecular stability.

\subsubsection{Molecules as Graphs} 
Molecules can also be represented as graphs, where atoms are nodes and bonds are edges. 
For MolGAN, molecules are undirected graphs $G$ with edges $E$ and nodes $V$, we employ the adjacency matrix formulation, \ref{app:math}. 
Each atom is represented by a one-hot vector, and each bond type is represented as an adjacency tensor. 
This graphical approach captures molecular connectivity directly, unlike 1D representations.

\subsection{MolGAN}
\label{molgan_section}

\begin{figure}[htb]
    \centering
    \includegraphics[width=\linewidth]{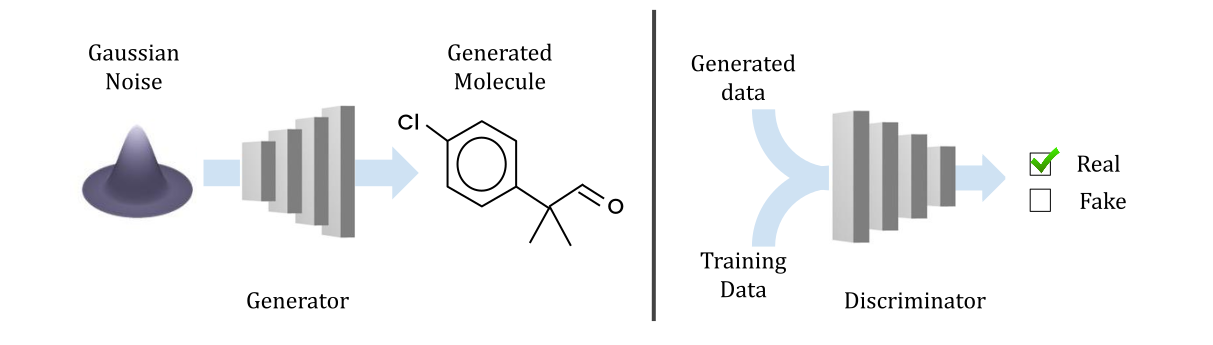}
    \caption{\textit{Model Architecture of MolGAN}}
    \label{fig:molgan_architecture}
\end{figure}

Molecular Generative Adversarial Network (MolGAN) represents a novel approach in the Generative Space for molecules by employing the GAN framework, which allows for an implicit, likelihood-free generative model which overcomes various problems (like graph matching~\citep{graph-matching} and node ordering heuristics) with previous models. 
MolGAN performs similarly to current SMILES-based approaches as well, albeit it is more susceptible to model collapse. 
We have incorporated the following loss used by the paper
\begin{equation}
        L(\mathbf{x}^{(i)}, G_{\theta}(z^{(i)});\phi) = \underbrace{-D_{\phi}(\mathbf{x}^{(i)}) + D_{\phi}(G_{\theta}(z^{(i)}))}_\text{Original WGAN Loss} 
        + \underbrace{\alpha(\lvert\lvert \nabla_{\hat{x}^{(i)}}D_{\phi}(\hat{x}^{(i)}) \rvert\rvert - 1)^2}_\text{Gradient Penalty}
\end{equation} 
\\[1mm]
This is an improved form of the WGAN~\citep{arjovsky2017wasserstein} loss, where $G_\theta$ is the generator, $D_\phi$ is the discriminator, $\mathbf{x}^{(i)} \sim p_{data}(\mathbf{x})$ and $\mathbf{z}^{(i)} \sim p_z(\mathbf{z})$ and $\mathbf{\hat{x}}^{(i)}$ is a sampled linear combination as the following equation $\mathbf{\hat{x}}^{(i)} = \epsilon \mathbf{x}^{(i)} + (1 - \epsilon)G_{\theta}(\mathbf{z}^{(i)})$ with $\epsilon \sim U(0, 1)$.
The model architecture is displayed in Figure \ref{fig:molgan_architecture}.


\subsection{Normalizing Flows}
\label{nflows_section}

\begin{figure}[!htb]
    \centering
    \includegraphics[width=\linewidth]{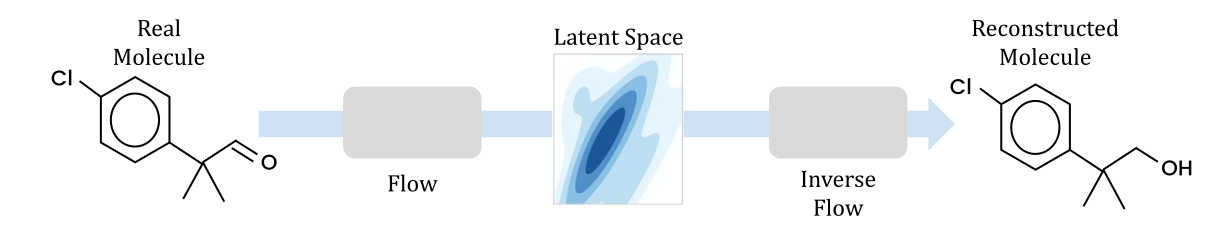}
    \caption{\textit{Model Architecture of Normalizing Flow models}}
    \label{fig:nflows_architecture}
\end{figure}

Normalizing Flows constitute a generative model that uses invertible transformations to model a probability distribution. 
This methodology enables the computation of likelihoods and the generation of samples by transforming simple base distributions (e.g., gaussian) into more complex ones through a sequence of invertible (\& differentiable) transformations.
Normalizing flows allow for direct sampling from the target distribution and thus are amazing for property-guided generation.
The model architecture is displayed in Figure \ref{fig:nflows_architecture}.

\section{Implementation}
\label{implementation}
The generative models were implemented in three main components: Layers, Base Model, and Molecule Generation Pipeline. 
We standardize (``deepchemize”~\citep{deepchem_alt}) molecule generation by providing \texttt{BasicMolGANModel} and \texttt{NormalizingFlowModel} which are highly flexible, allowing users to experiment with generators, discriminators, flow layers, and more.

\subsection{Layers}
\textbf{MolGAN}: The Discriminator in MolGAN begins with an encoder layer composed of multiple convolutional layers, followed by an aggregation layer. The architecture is flexible, allowing easy customization based on user requirements.

\textbf{Normalizing Flows}: Normalizing Flow layers are responsible for both forward and backward computation. In DeepChem, the Normalizing Flows pipeline supports linear layers, which are computationally efficient compared to more complex layers like planar or autoregressive. These layers use linear transformations to model relationships between molecular dimensions:
$
\mathbf{g(x)} = \mathbf{Wx} + \mathbf{b}
$
Here, $W \in \mathbb{R}^{D \times D}$ and $b \in \mathbb{R}^D$ are parameters, and if $W$ is invertible, the function is invertible as well.

\subsection{Base Model}
The Base Model consists of three primary components: the Generator, Discriminator, and Normalizing Flow model. Each component is modular and can be adjusted independently for flexibility in experimentation.

The \textbf{Generator} is modeled as an MLP~\citep{MURTAGH1991183} with varying units to implicitly model a probability distribution over molecular graphs. It outputs continuous objects representing nodes and edges, which are transformed using the Gumbel Softmax trick~\citep{gumbel}. The model can also be adapted to use other methods such as the straight-through estimator~\citep{straight-through}. 

The \textbf{Discriminator} scores the generated molecular graph through a series of relational graph convolutional layers~\citep{rgcn}. These layers update node representations based on their neighbors and edge types. After several layers of convolution, MLPs extract the final score of the graph. More detailed mathematical formulations of the Generator, Discriminator, and Normalizing Flow model are mentioned in Appendix \ref{app:math}

\subsection{Molecule Generation Pipelines}

\subsubsection{MolGAN}
\begin{figure*}[h]
    \centering
    \includegraphics[width=1\linewidth]{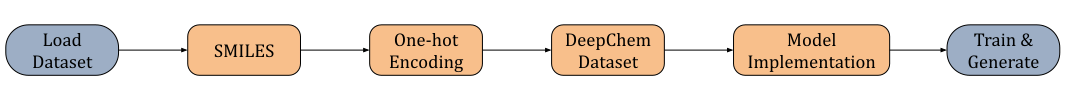}
    \caption{\textit{Molecule generation pipeline for MolGAN}}
    \label{fig:molgan-fig}
\end{figure*}
Training / Generating from MolGAN on custom datasets follows a very simple pipeline, which can be achieved in a few lines of code; we describe the pipeline in Figure \ref{fig:molgan-fig}, which involves extracting SMILES~\citep{smiles} representations and wrapping them in a DeepChem dataset that can be used to train the model.
The full pipeline is demonstrated in \ref{app:molgan-code} with a few lines of code.

\subsubsection{Normalizing Flows}
\begin{figure*}[h]
    \centering
    \includegraphics[width=1\linewidth]{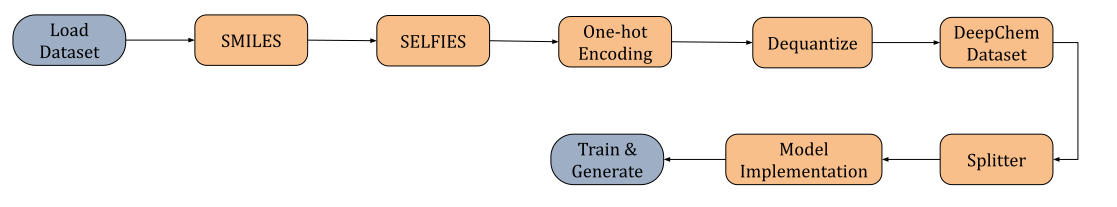}
    \caption{\textit{Molecule generation pipeline for Normalizing flows}}
    \label{fig:norm-fig}
\end{figure*}
Analogous to the MolGAN pipeline, the Normalizing Flows pipeline (Figure \ref{fig:norm-fig}) expects a SELFIES~\citep{selfies} string as input. 
There are added steps for Dequantization (i.e., adding noise in [0, 1) to every input)
The full pipeline is demonstrated in \ref{app:norm-flow-code} with a few lines of code.

\section{Experiments}

\subsection{Datasets}
We use the following publicly available datasets: QM7~\citep{qm7}, BBBP~\citep{bbbp}, Lipophilicity~\citep{lipo}, PPB~\citep{ppb}, and QM9~\citep{qm9}. 
QM7 is a subset of GDB-13, consisting of molecules with up to 7 heavy atoms (C, N, O, and S). The BBBP dataset includes around 2,000 molecules with binary labels related to blood-brain barrier permeability. The Lipophilicity dataset, sourced from ChEMBL, contains 4,200 compounds. The PPB dataset, curated from PubChem BioAssay by the Maximum Unbiased Validation (MUV) group, includes around 11,000 compounds. QM9 is a subset of GDB-17, comprising 134,000 organic molecules with up to 9 heavy atoms. All datasets were obtained through MoleculeNet~\citep{moleculenet}.

For nomalizing flows we utilize the full dataset for all experiments, however, following~\citep{molgan}, for MolGAN, we only use molecules that only have C, N, O and F in them.
The final number of valid molecules in every dataset are shown in \ref{tab:datasets}.
\begin{table}[h!]
    \centering
    \begin{tabular}{@{}ll@{}}
    \toprule
    \textbf{Dataset}       & \textbf{Number of Samples} \\ \midrule
    BBBP                  & 1631 \\
    PPB                   & 1291 \\
    QM7                   & 5470 \\
    QM9                   & 105984 \\
    Lipophilicity         & 3360 \\ \bottomrule
    \end{tabular}
    \vspace{0.5cm}
    \caption{Number of samples in each dataset.}
    \label{tab:datasets}
\end{table}

\subsection{Experimental Setup} 
\label{expts}
For \textbf{MolGAN}, we use four edge types (single, double, triple, and no-bond), five node types (C, N, O, F, and PAD) similar to~\citep{molgan}, whilst our discriminator discriminates among molecules of all lengths, our generator is restricted to small molecule generation, limited to 9 vertices.
To allow this, we would set the maximum number of allowed atoms to the maximum in a dataset.
We report mean results across 10 seeds, each involving generation of 6400 molecules, as in~\citep{molgan}.

For \textbf{Normalizing Flows}, we used two layers of flows with Masked Affine Flows~\citep{maf} along with an ActNorm layer~\citep{glow}. 
We used a Multivariate Normal Distribution from PyTorch~\citep{pytorch} to build our flow model. 
All hyper-parameters are listed in \ref{app:hyperparams}.
All experiments used Python 3.10.

\subsection{Results}

\begin{table}[h]
    \centering
    \begin{tabular}{l l S[table-format=3.2] S[table-format=3.2] S[table-format=3.2] S[table-format=2.2] S[table-format=2.2]}
        \toprule
        \textbf{Dataset} & \textbf{Model} & \textbf{Val($\uparrow$)} & \textbf{Uni($\uparrow$)} & \textbf{Nov($\uparrow$)} & \textbf{SAS($\downarrow$)} & \textbf{Dru($\uparrow$)} \\
        \midrule
        \multirow{2}{*}{QM7} & MolGAN & 92.1 & 4.18 & 100.0 & 2.11 & 70.45 \\
                             & Norm Flow & 100.0 & 91.87 & 100.0 & 5.29 & 52.19 \\
        \cmidrule{1-7}
        \multirow{3}{*}{QM9} & MolGAN & 89.67 & 4.46 & 100.0 & 2.83 & 66.03 \\
                             & Norm Flow & 100.0 & 99.15 & 100.0 & 6.58 & 44.55 \\
                             & Original Implementation~\citep{molgan} & 87.7 & 2.9 & 97.7 & \longdash* & \longdash* \\
        \cmidrule{1-7}
        \multirow{2}{*}{BBBP} & MolGAN & 57.18 & 0.28 & 100.0 & 5.24 & 55.42 \\
                              & Norm Flow & 92.83 & 98.98 & 100.0 & 5.76 & 49.50 \\
        \cmidrule{1-7}
        \multirow{2}{*}{Lipophilicity} & MolGAN & 49.57 & 0.08 & 100.0 & 5.75 & 53.58 \\
                                       & Norm Flow & 100.0 & 99.36 & 100.0 & 6.18 & 49.39 \\
        \cmidrule{1-7}
        \multirow{2}{*}{PPB} & MolGAN & 14.04 & 25.07 & 100.0 & 9.17 & 18.31 \\
                                  & Norm Flow & 100.0 & 98.60 & 100.0 & 5.95 & 50.30 \\
        \bottomrule
    \end{tabular}
    \vspace{0.25cm}
    \caption{
    Evaluation performance of the open-source models on QM7, BBBP, Lipophilicity, PPB, QM9 datasets. The models are evaluated on the Validity of molecules generated (Val), Uniqueness (Uni), Novelty (Nov), Synthetic Accessibility Score (SAS) out of 10, and Druglikeliness (Dru). SAS is reported out of 10, whereas everything else is a \%. (* - not reported in the paper)
    }
    \label{tab:res}
\end{table}
Our results \ref{tab:res} demonstrate that our implementation's performance is comparable to existing work (~\citep{molgan, molgrow}).
For a fair comparison, we only compare our implementation to the non-RL results from~\citep{molgan}, also~\citep{molgan} only train and evaluate on QM9.

\section{Conclusion \& Discussion}
In this work, we improve DeepChem's generative modeling tools and provide a more standardized and scalable implementation that makes generative molecular methods more accessible to scientists. 
Standard Machine Learning practices are built within DeepChem (e.g., checkpointing, validation, logging, etc.), which would otherwise need some form of human expertise. 
Benchmarks show comparable performance with existing implementations, and the tight integration with DeepChem facilitates fast future improvements. 
This will also allow for Reinforcement Learning~\citep{molgan} based approaches to be incorporated easily. 
We observed high variance for datasets where the average number of atoms in a molecule was much larger than 9, and leave this for future work.
We will provide multi-GPU training support in future work using Distributed Data Parallelism~\citep{ddp}, Sharding \& PyTorch~\citep{pytorch}.

\section*{Impact Statement}

This paper makes Generative Molecular modeling accessible to a broad spectrum of people, and while this is intended, it is not tough to modify these models to make them generate toxic and harmful molecules. 
However, the synthesis of novel molecules is an area that requires a lot of human expertise, and in general, is not easy to do. 
In a future where synthesis methods are also equally accessible, these models can be potentially dangerous, but at the present time, the positive outcomes outweigh the negative ones.

\bibliographystyle{plain}
\bibliography{main.bib}


\section*{Appendix}

\appendix




\section{Mathematical Formulations}
\label{app:math}
\subsection{Molecules as Graphs}
For MolGAN, each molecule can be represented as an undirected Graph $G$ with a set of edges $E$ and nodes $V$. 
Each atom $v_i \in V$ is associated with a $D$-dimensional one-hot vector $\mathbf{x}_i$. 
Each edge $(v_i, v_j) \in E$ is also associated with a bond type $y \in \{1, \dots, Y\}$. 
Thus, we have a representation of a graph as two objects: a $\mathbf{X} = [\mathbf{x}_1, \dots, \mathbf{x}_n]^T \in \mathbb{R}^{N \times D}$ and an adjacency tensor $\mathbf{A} \in \mathbb{R}^{N \times N \times Y}$. $\mathbf{A}_{i, j} \in \mathbb{R}^Y$ is a one-hot vector indicating the type of edge between $i$ and $j$.
\subsection{Generator}
For any latent variable 
\begin{equation*}
    z \sim \mathcal{N}(\mathbf{0, I})
\end{equation*}
the generator $\mathcal{G}_{\theta}(z)$ outputs two continuous objects:
\begin{equation*}
    \mathbf{X} \in \mathbb{R}^{N \times D}, \mathbf{A} \in \mathbb{R}^{N \times N \times D}
\end{equation*}
Using the Gumbel Softmax trick~\citep{gumbel}, categorical sampling is performed as:
\begin{align}
    \mathbf{\Tilde{X}} & = \mathbf{X} + \text{Gumbel}(\mu = 0, \beta = 1) \\
    \mathbf{\Tilde{A}} & = \mathbf{A}_{ijy} + \text{Gumbel}(\mu = 0, \beta = 1)
\end{align}
Alternatively, $\mathbf{\Tilde{X}}$ and $\mathbf{\Tilde{A}}$ can remain as $\mathbf{X}$ and $\mathbf{A}$.

\subsection{Discriminator}
The discriminator updates node representations through relational graph convolution as:
\begin{align}
    \boldsymbol{h}_i^{\prime(\ell+1)} & = f_s^{(\ell)}\left(\boldsymbol{h}_i^{(\ell)}, \boldsymbol{x}_i\right) + \sum_{j=1}^N \sum_{y=1}^Y \frac{\tilde{\boldsymbol{A}}_{i j y}}{\left|\mathcal{N}_i\right|} f_y^{(\ell)}\left(\boldsymbol{h}_j^{(\ell)}, \boldsymbol{x}_j\right) \\
    \boldsymbol{h}_i^{(\ell+1)} & = \tanh \left(\boldsymbol{h}_i^{\prime(\ell+1)}\right)
\end{align}
where $\boldsymbol{h}_i^{(l)}$ represents the signal at node $i$ and layer $l$, and $f_s^{(l)}$ is a self-connection transformation. After several layers of graph convolution, the graph score is calculated as:
\begin{align}
\boldsymbol{h}_{\mathcal{G}}^{\prime} & = \sum_{v \in \mathcal{V}} \sigma\left(i\left(\boldsymbol{h}_v^{(L)}, \boldsymbol{x}_v\right)\right) \odot \tanh \left(j\left(\boldsymbol{h}_v^{(L)}, \boldsymbol{x}_v\right)\right) \\
\boldsymbol{h}_{\mathcal{G}} & = \tanh(\boldsymbol{h}_{\mathcal{G}}^{\prime})
\end{align}
where $\sigma$ is the sigmoid function $ \sigma = \frac{1}{1 + e^{-x}}$.

\subsection{Normalizing Flow Model}
The Normalizing Flow model starts with a base distribution $p_z$ and transforms it into a density function $p_x$. The transformation $x = L(z)$ is invertible, and the probability density is updated as:
\begin{equation*}
    \rho_X(x) = \rho_Z(z) \cdot |det(J_L(z))|^{-1}
\end{equation*}
where $J_L(z)$ is the Jacobian determinant of the transformation.

\section{Code Snippets}
\subsection{MolGAN Pipeline}
\label{app:molgan-code}
Training -
\begin{lstlisting}[language=Python]
    from deepchem.models.torch_models import BasicMolGANModel as MolGAN
    from deepchem.models.optimizers import ExponentialDecay
    import deepchem as dc
    import torch.nn.functional as F
    import torch
    ...
    gan = MolGAN(learning_rate=ExponentialDecay(0.001, 0.9, 5000))
    dataset = dc.data.NumpyDataset([x.adjacency_matrix for x in features],[x.node_features for x in features])
    def iterbatches(epochs):
        for i in range(epochs):
            for batch in dataset.iterbatches(batch_size=gan.batch_size, pad_batches=True):
                adjacency_tensor = F.one_hot(
                        torch.Tensor(batch[0]).to(torch.int64),
                        gan.edges).to(torch.float32)
                node_tensor = F.one_hot(
                        torch.Tensor(batch[1]).to(torch.int64),
                        gan.nodes).to(torch.float32)
                yield {gan.data_inputs[0]: adjacency_tensor, gan.data_inputs[1]:node_tensor}
    # train model
    gan.fit_gan(iterbatches(8), generator_steps=0.2, checkpoint_interval=0)
\end{lstlisting}
Inference - 
\begin{lstlisting}[language=Python]
    generated_data = gan.predict_gan_generator(10)
    # convert graphs to RDKitmolecules
    new_mols = feat.defeaturize(generated_data)
\end{lstlisting}

\subsection{NormalizngFlows Pipeline}
\label{app:norm-flow-code}
Training - 
\begin{lstlisting}[language=Python]
    import deepchem as dc
    from deepchem.data import NumpyDataset
    from torch.distributions.multivariate_normal import MultivariateNormal
    from rdkit import Chem
    from dc.torch_models.nflows import *
    
    # Pass data through pipeline (mentioned above)
    ...
    
    # Construct flow model
    flows = [ActNorm(latent_size)]
    nfm = NormFlow(flows, distribution, dim)
    nfm.fit(max_iterations, optimizer)
\end{lstlisting}
Inference -
\begin{lstlisting}[language=Python]
    mols = nfm.generate(num_molecules=1000)
    valid_mols = validate_mols(mols)
\end{lstlisting}

\section{Hyperparameters}
\label{app:hyperparams}

For our MolGAN implementation, similar to~\citep{molgan}, we vary the given hyperparameters, and pick the best for each dataset prioritising Validity of generations.
Except for QM9, we train each model for 30 epochs on the whole dataset.
QM9 is trained similar to ~\citep{molgan}, i.e., 300 epochs on a random 5k subset, we also perform early stopping if the uniqueness goes below 2\%.
Apart from this we fix Generator Steps to 0.2, i.e., one step every 5 discriminator steps, a batch size of 32, learning rate of $1e-4$, and an embedding dimension of 32.
\begin{table}[h!]
\centering
\begin{tabular}{@{}ll@{}}
\toprule
\textbf{Hyperparameter} & \textbf{Values} \\ \midrule
Dropout Rates           & $\{0.0, 0.1, 0.25\}$ \\
Sampling Mode           & Straight-through, Gumbel, SoftMax \\ \bottomrule
\end{tabular}
\vspace{0.5cm}
\caption{Hyperparameter settings for MolGAN.}
\label{tab:hyperparameters}
\end{table}

For the Normalizing Flows implementation, we use a learning rate of 1e-4, and an equal weight decay.
We train on a batch size of 1024 for 100 epochs.
Both the models were trained on with Adam optimizer~\citep{adam}.

\end{document}